\documentclass[runningheads]{llncs}

\usepackage[T1]{fontenc}
\usepackage{graphicx}
\usepackage{tabularx}
\usepackage{multirow}
\usepackage{amsmath}
\usepackage{amsfonts}

\newcolumntype{Y}{>{\centering\arraybackslash}X}

\begin{document}

\title{Federated Markov Imputation: Privacy-Preserving Temporal Imputation in Multi-Centric ICU Environments}

\titlerunning{Federated Markov Imputation}

\author{Christoph Düsing\orcidID{0000-0002-7817-9448} \and
Philipp Cimiano\orcidID{000-0002-4771-441X}}
\authorrunning{C. Düsing \& P. Cimiano}
\institute{CITEC, Bielefeld University, Bielefeld, Germany \\ 
\email{\{cduesing, cimiano\}}@techfak.uni-bielefeld.de}

\maketitle  

\begin{abstract}
Missing data is a persistent challenge in federated learning on electronic health records, particularly when institutions collect time-series data at varying temporal granularities. To address this, we propose \textit{Federated Markov Imputation} (FMI), a privacy-preserving method that enables \textit{Intensive Care Units} (ICUs) to collaboratively build global transition models for temporal imputation. We evaluate FMI on a real-world sepsis onset prediction task using the MIMIC-IV dataset and show that it outperforms local imputation baselines, especially in scenarios with irregular sampling intervals across ICUs.

\end{abstract}

\section{Introduction}

\textit{Federated Learning} (FL) has emerged as a privacy-preserving machine learning paradigm that allows multiple hospitals or \textit{Intensive Care Units} (ICUs) to collaboratively train models without exchanging raw patient data~\cite{mcmahan2017communication}. This makes FL particularly attractive for high-stakes medical tasks such as predicting sepsis onset~\cite{dusing2024integrating,mondrejevski2023predicting}. However, many clinical prediction tasks -- including sepsis onset prediction -- rely on time-series data, often incomplete due to irregular sampling or clinical priorities. In FL, the challenge of missing data is amplified because participants must agree on a uniform data format and time intervals. If hospitals collect data at different temporal granularities, coarser institutions must impute more data to align with the global format. This restricts the applicability of local imputation techniques, especially those relying on fine-grained temporal transitions, and necessitates the development of FL-specific imputation methods.

\noindent\textbf{Related Work.}
While federated data imputation gained some attention in recent years, research in this area remains scarce. Initial approaches such as \textit{Cafe}~\cite{min2025cafe} propose federated imputation by modeling heterogeneity in missing data across silos. \textit{FedBeat}~\cite{wang2024federated} introduces transition-matrix-based modeling for noisy label correction but focuses on label noise rather than temporal features. Similarly, Gkillas et al \cite{gkillas2022missing} propose federated time-series imputation for industrial settings using an jointly trained autoencoder. However, no work has yet considered the construction of \textit{Federated Markov Chains} for imputation in time-series clinical FL settings, especially when facing different temporal granularities.

\noindent\textbf{Contributions.}
To this end, we make the following contributions: (1) we propose \textit{Federated Markov Imputation} (FMI), a privacy-preserving method for time-series data imputation in FL; (2) we apply FMI to a real-world sepsis prediction task under both uniform and heterogeneous temporal granularities; and (3) we show that FMI improves the predictive performance over two baselines, especially when facing heterogeneous time intervals among participating ICUs.

\section{Federated Markov Imputation}

The proposed method for FMI consists of three steps outlined in the following: 

\noindent\textbf{\textit{Step 1: Local Transition Matrix.}}  
Each ICU discretizes its local time-series data into $n$ bins per feature and constructs a first-order Markov transition matrix $T_c \in \mathbb{R}^{n \times n}$, where $T_c(i, j)$ denotes the empirical probability of transitioning from bin $i$ to $j$ between adjacent time steps. Transitions are computed only from observed, non-missing data and remain local to the ICU.

\noindent\textbf{\textit{Step 2: Federated Transition Matrix.}}  
Using \textit{Secure Aggregation}~\cite{bonawitz2017practical}, all ICUs contribute their local transition counts, which are masked and summed to compute a global transition matrix $T_{\text{fed}}$. The aggregation is performed without exposing individual ICU statistics, ensuring data privacy.

\noindent\textbf{\textit{Step 3: Federated Markov Imputation.}}  
For each missing value, the ICU imputes the most likely bin using $T_{fed}$ and the available temporal context. If both the preceding bin $b_{t-1}$ and succeeding bin $b_{t+1}$ are known, the missing bin $b_t$ is selected as:
{\small
    \begin{equation}
        \hat{b}_t = \arg\max_j \; T_{\text{fed}}(b_{t-1}, j) \cdot T_{\text{fed}}(j, b_{t+1})
    \end{equation}
}

This selects the bin that best fits the expected transitions on both sides. If only one neighbor is known, a single-directional transition is used. For consecutive missing values, the most probable path is inferred recursively. The final imputed value corresponds to the midpoint of the selected bin $\hat{b}_t$.

\section{Preliminary Evaluation}

\subsection{Experimental Setup}

\noindent\textbf{Data Setting.}
This study utilized the \textit{MIMIC-IV} dataset \cite{mimic}, containing both septic and non-septic patients admitted to one of seven ICUs, namely \textit{Medical ICU} (MICU), \textit{Medical/Surgical ICU} (MICU/SICU), \textit{Surgical ICU} (SICU), \textit{Trauma SICU} (TSICU), \textit{Coronary Care Unit} (CCU), \textit{Cardiac Vascular ICU} (CVICU), and \textit{Neuro SICU} (NSICU).
With respect to data acquisition and distribution, we follow the work of Düsing and Cimiano \cite{dusing2025shapfl}. Specifically, we identify 28,610 patients relevant to the study and assign each patient to the ICU they were initially admitted to. Per patient, we identify 25 clinically relevant features, which are aggregated into six 1h windows following admission. The sepsis labels are assigned following the \textit{Sepsis-3} definition, where each patient is labeled as positive if they developed sepsis within the first 30h of ICU admission.
\\
During evaluation, we consider two temporal settings. In the \textit{Regular} setting, all ICUs retain the hourly data. To simulate heterogeneous sampling in the \textit{Irregular} setting, we randomly assign two ICUs to 2-hour and two to 3-hour intervals, removing intermediate values accordingly. This setup allows us to assess imputation performance under varying temporal granularities across ICUs.

\noindent\textbf{Model Setting.}
In line with similar studies \cite{dusing2025shapfl,mondrejevski2023predicting}, we use an model consisting of an input layer, followed by three LSTM layers with 16 units each. Each LSTM layer includes batch normalization and a dropout of 0.2 to prevent overfitting. A subsequent linear layer reduces the dimensionality to 8 units, before a final linear layer provides a binary prediction indicating the likelihood of sepsis onset.
\\
The model is trained for 50 rounds using \textit{FedAvg} \cite{mcmahan2017communication} as aggregation strategy.

\noindent\textbf{Evaluation Setting.}
We compare our method against two baselines: (1) \textit{local mean} imputation, where missing values are filled using each ICU’s feature-wise mean, and (2) \textit{Local Markov Imputation} (LMI), which applies Markov-based imputation using each ICU’s local transition matrix $T_c$ only. However, in the \textit{Irregular} setting, LMI becomes infeasible for ICUs with 2h or 3h intervals, as they lack the hourly resolution needed for the federated training.
\\
Performance is measured using the $AUC$, which reflects the classifier’s ability to distinguish between septic and non-septic cases across different thresholds.

\subsection{Performance Analysis}

Table~\ref{table:evaluation} presents the $AUC$ scores across all ICUs. In the \textit{Regular} setting, the main findings are: 
(1) FMI improves the mean AUC over both baselines, though the gains remain moderate;  
(2) FMI yields better performance than LMI for 5 out of 7 ICUs, while two ICUs benefit slightly more from using LMI; 
(3) all three imputation modes produce federated models with $AUC$ > 0.8, generally considered the threshold for clinical applicability~\cite{dusing2025shapfl}.
\\
In the \textit{Irregular} setting, we observe the following:  
(1) overall performance declines compared to the \textit{Regular} setting, which is expected due to increased scarcity and lower data resolution;  
(2) the performance of local mean imputation degrades more significantly, with the federated model falling below the clinical $AUC$ threshold of 0.8;  
(3) performance becomes more uneven across ICUs --those with 1h intervals achieve similar or even better results than in the \textit{Regular} setting, while those with 2h or 3h intervals show noticeable declines;  
(4) FMI reduces the negative impact of irregular sampling, especially for MICU/SICU and NSICU (3h intervals), where improvements are particularly pronounced.

\begin{table}[t]
    \begin{center}
    \caption{$AUC$ for Sepsis Onset Prediction Using Different Imputation Modes Applied to Regular and Irregular Time Intervals (Best Performance in Bold and Underlined)}
    \label{table:evaluation}
    \fontsize{8}{10}\selectfont
    \resizebox{1\textwidth}{!}{
        \begin{tabularx}{\textwidth}{c|c||Y|Y|Y|Y|Y|Y|Y||Y}
            Time & Imputation & \multirow{2}{*}{MICU} & MICU/ & \multirow{2}{*}{SICU} & \multirow{2}{*}{TSICU} & \multirow{2}{*}{CCU} & \multirow{2}{*}{CVICU} & \multirow{2}{*}{NSICU} & \multirow{2}{*}{Mean} \\
            Intervals & Mode &  & SICU &  &  &  &  &  &  \\
            \hline\hline
            \multirow{6}{*}{\rotatebox[origin=c]{90}{Regular}}& Local & 0.8520 & 0.8696 & 0.8931 & 0.9040 & 0.8549 & 0.5400 & 0.8454 & 0.8634\\
            & Mean & \tiny{±0.00} & \tiny{±0.00} & \tiny{±0.00} & \tiny{±0.00} & \tiny{±0.00} & \tiny{±0.00} & \tiny{±0.00} & \tiny{±0.03}\\
            \cline{2-10}
            &\multirow{2}{*}{LMI} & 0.8639 & \textbf{\underline{0.8992}} & 0.9039 & 0.8957 & 0.7986 & \textbf{\underline{0.8250}} & 0.8609 & 0.8710\\
            & & \tiny{±0.00} & \tiny{±0.00} & \tiny{±0.00} & \tiny{±0.00} & \tiny{±0.00} & \tiny{±0.00} & \tiny{±0.00} & \tiny{±0.01}\\
            \cline{2-10}
            & FMI  & \textbf{\underline{0.8648}} & 0.8950 & \textbf{\underline{0.9074}} & \textbf{\underline{0.9162}} & \textbf{\underline{0.8314}} & 0.8247 & \textbf{\underline{0.8818}} & \textbf{\underline{0.8878}}\\
            & \tiny{(ours)} & \tiny{±0.00} & \tiny{±0.00} & \tiny{±0.00} & \tiny{±0.00} & \tiny{±0.00} & \tiny{±0.00} & \tiny{±0.00} & \tiny{±0.01} \\
            \hline
            \hline
            \multirow{7}{*}{\rotatebox[origin=c]{90}{Irregular}} & Interval (h) & 1h & 3h & 2h & 1h & 2h & 1h & 3h & n/a \\
            \cline{2-10}
            & Local  & 0.8738 & 0.5958 & 0.8607 & 0.8967 & 0.8180 & 0.7538 & 0.5180 & 0.7961\\
            & Mean & \tiny{±0.00} & \tiny{±0.00} & \tiny{±0.00} & \tiny{±0.00} & \tiny{±0.00} & \tiny{±0.00} & \tiny{±0.00} & \tiny{±0.05} \\
            \cline{2-10}
            & \multirow{2}{*}{LMI} & \multirow{2}{*}{n/a} & \multirow{2}{*}{n/a} & \multirow{2}{*}{n/a} & \multirow{2}{*}{n/a} & \multirow{2}{*}{n/a} & \multirow{2}{*}{n/a} & \multirow{2}{*}{n/a} & \multirow{2}{*}{n/a}\\
            & & & & & & & & &\\
            \cline{2-10}
            & FMI  & \textbf{\underline{0.9196}} & \textbf{\underline{0.8038}} & \textbf{\underline{0.8845}} & \textbf{\underline{0.9198}} & \textbf{\underline{0.8452}} & \textbf{\underline{0.7777}} & \textbf{\underline{0.8503}} & \textbf{\underline{0.8629}}\\
            & \tiny{(ours)} & \tiny{±0.00} & \tiny{±0.00} & \tiny{±0.00} & \tiny{±0.00} & \tiny{±0.00} & \tiny{±0.00} & \tiny{±0.00} & \tiny{±0.01}\\
        \end{tabularx}
    }
    \end{center}
\end{table}

\section{Conclusion and Outlook}

We introduced FMI, a privacy-preserving method for time-series imputation in federated clinical settings. Our preliminary results on sepsis prediction show that FMI improves performance over both baselines, particularly under irregular sampling. These findings highlight the potential of FMI to harmonize heterogeneous clinical data and impute missing values without compromising privacy.
\\
In future work, we aim to conduct more detailed analyses, extend comparisons with additional baselines, and validate FMI in a real-world clinical case-study.

\end{document}